\documentclass{article}
\usepackage{iclr2027_conference,times}
\usepackage[T1]{fontenc}
\usepackage{microtype}
\usepackage{amsmath,amssymb}
\usepackage{booktabs}
\usepackage{graphicx}
\usepackage{xcolor}
\usepackage{url}
\usepackage{pgfplots}
\pgfplotsset{compat=1.18}
\usetikzlibrary{arrows.meta,positioning}
\usepgfplotslibrary{groupplots}
\usepackage{multirow,longtable}
\usepackage{float}
\usepackage[hidelinks]{hyperref}
\usepackage{tikz}
\usetikzlibrary{positioning,arrows.meta,fit,backgrounds}
\usepackage{wrapfig}
\usepackage{booktabs}
\usepackage[table]{xcolor}

\newcommand{\evA}{\texttt{evA}}

\title{Learning a Fact Is Not Learning How to \\Retrieve It}

\author{Chaemin Jang, Jihee Kim$^{\dagger}$, Dongman Lee$^{\dagger}$ \\
Korea Advanced Institute of Science and Technology (KAIST) \\
\texttt{jchaemin@kaist.ac.kr}
}
\iclrfinalcopy

\date{}
\begin{document}

\begingroup
\renewcommand{\thefootnote}{$\dagger$}
\footnotetext{Co-corresponding authors.}
\endgroup

\maketitle

\begin{abstract}
A model trained on {\small\texttt{The capital of X is Y}} may produce {\small\texttt{Y}} after {\small\texttt{The capital of X is}} but fail after {\small\texttt{The capital of X:}}. We call these different ways of eliciting the same fact \emph{request forms}. To separate learning a fact from retrieving it, we train two models in two stages. In the first stage (request-form training), one model sees each fact in five forms and the other sees the same facts only as statements. In the second stage (target-fact training), both receive identical training on new facts, all as statements. Both then retrieve the new facts almost equally well from statements, but differ sharply on other request forms. Thus, a model can learn how to retrieve through a request form before it learns the facts. To understand this difference, we examine the hidden state immediately before the answer, which we call the \emph{context state}. When given two different request forms for the same fact, the model trained on five forms in stage one produces more similar context states than the model trained on statements alone in that stage. Changing this state at retrieval time can enable or prevent retrieval of an already learned fact, and the same effect transfers across facts and factual relations, such as capitals and currencies. To test its role during learning, we change the context state only during target-fact training. This intervention changes later retrieval without intervention at test time. Together, these results show that later retrieval depends on earlier request-form experience and the context state during fact learning.
%A model may learn ``The capital of X is Y'' but fail when asked ``Capital of X:''. We call these different ways of asking for the same fact \emph{request forms}. We separate learning a fact from learning how to retrieve it by first training models on different request forms, then giving them exactly the same training on the same new facts. The models learn the new facts equally well in the form used for training, yet differ sharply in which other request forms can retrieve them. Thus, a model can learn how to retrieve through a request form before it learns the facts. We find that retrieval is controlled by the hidden state immediately before the answer, which we call the \emph{context state}. Changing this state can enable or prevent retrieval of an already learned fact, and the same effect transfers across facts and relations. Earlier request-form training changes the context states produced by those request forms. Most importantly, changing the context state while a new fact is being learned changes how that fact is retrieved later. Together, these results identify the context state as the link between earlier request-form learning and the later retrieval of new facts.
\end{abstract}

\section{Introduction}
Suppose we fine-tune a language model on {\small\texttt{The capital of X is Y.}} Later, the model produces {\small\texttt{Y}} after {\small\texttt{The capital of X is}} but fails after {\small\texttt{- Capital of X:}}. The fact can be retrieved through one form but not the other. We call these different ways of eliciting the same fact \emph{request forms}. Prior work has observed related gaps between learning facts from text and retrieving them in a new form \citep{allenzhu2023physics,jiang2024instruction}. But such a failure leaves a basic ambiguity: did the model learn the fact only weakly, or did it fail to learn how to retrieve it through the second request form? In ordinary fine-tuning, the fact and the form in which it is presented are learned together, making these explanations difficult to distinguish.

We separate these possibilities with a two-stage experiment. In the first stage (request-form training), one model sees facts only as statements such as {\small\texttt{The capital of A is B.}}, while the other sees those same facts as statements and in four additional forms, including {\small\texttt{Capital of A: B.}} and {\small\texttt{Name the capital of A: B.}}. In the second stage (target-fact training), both models receive identical training on new facts, all written as statements such as {\small\texttt{The capital of X is Y.}}. In the capitals-only experiment, both models produce {\small\texttt{Y}} after {\small\texttt{The capital of X is}} for nearly every evaluated fact. But after {\small\texttt{- Capital of X:}}, the first model almost never produces {\small\texttt{Y}}, while the second does so for 97.7\% of the evaluated facts. Because their target-fact training is identical, the retrieval difference comes from what they learned before X and Y were introduced.

%We separate these possibilities with a two-stage experiment. First, one model sees facts only as statements such as ``The capital of A is B,'' while another also sees facts written in different request forms such as ``Capital of A: B'' or ``Name the capital of A: B.'' Next, we train both models in exactly the same way on new facts: ``The capital of X is Y.'' Later, when we ask both models ``The capital of X is'', they behave almost identically: both produce ``Y'' nearly every time. But when we ask ``- Capital of X:'', the first model almost never produces ``Y'', while the second does so 97.7\% of the time. Since both models received exactly the same training on X and Y, the difference must come from what they learned before X and Y were introduced.

To understand this difference, we examine the hidden state immediately before the answer, which we call the \emph{context state}. Replacing this state at retrieval time can reverse the result: a state from a successful request can make a failing request succeed, while a state from the failing request can make a successful one fail. A context state from one fact can also help retrieve another fact, even from a different factual relation. This suggests that part of its effect on retrieval can be reused across facts.

We next ask whether this state also connects earlier request-form training to later fact learning. For the same new fact, the declarative and list request forms produce more similar context states in the model trained on five forms in stage one than in the model trained only on statements at that stage. To test whether the state present during learning matters, we train a model on the same factual sentences as before but shift its context state toward one produced by a different request form. When training ends, we stop making the shift and test normally. The facts can now be retrieved through a request form that previously failed, without intervention at test time. Thus, how a fact can later be retrieved is shaped in part by the context state present while it was learned.

Our contributions are:
\begin{itemize}
    \item We separate learning a fact from learning how to retrieve it.
    \item We identify the \emph{context state} as the link between what the model learns from earlier request forms and how new facts are later retrieved, and show that it can control retrieval across facts and relations.
    \item We show that changing the context state during factual fine-tuning changes how the learned fact is later retrieved.
\end{itemize}

%Our contributions are:
%\begin{itemize}
%    \item We separate learning a fact from learning how to retrieve it. 
%    \item We identify the \emph{context state} as the link between what the model learns from earlier request forms and how new facts are later retrieved.
%    \item We show that changing the context state during factual fine-tuning changes how the learned fact is later retrieved.
%\end{itemize}

\begin{figure*}[t]
    \centering
    \includegraphics[width=\textwidth]{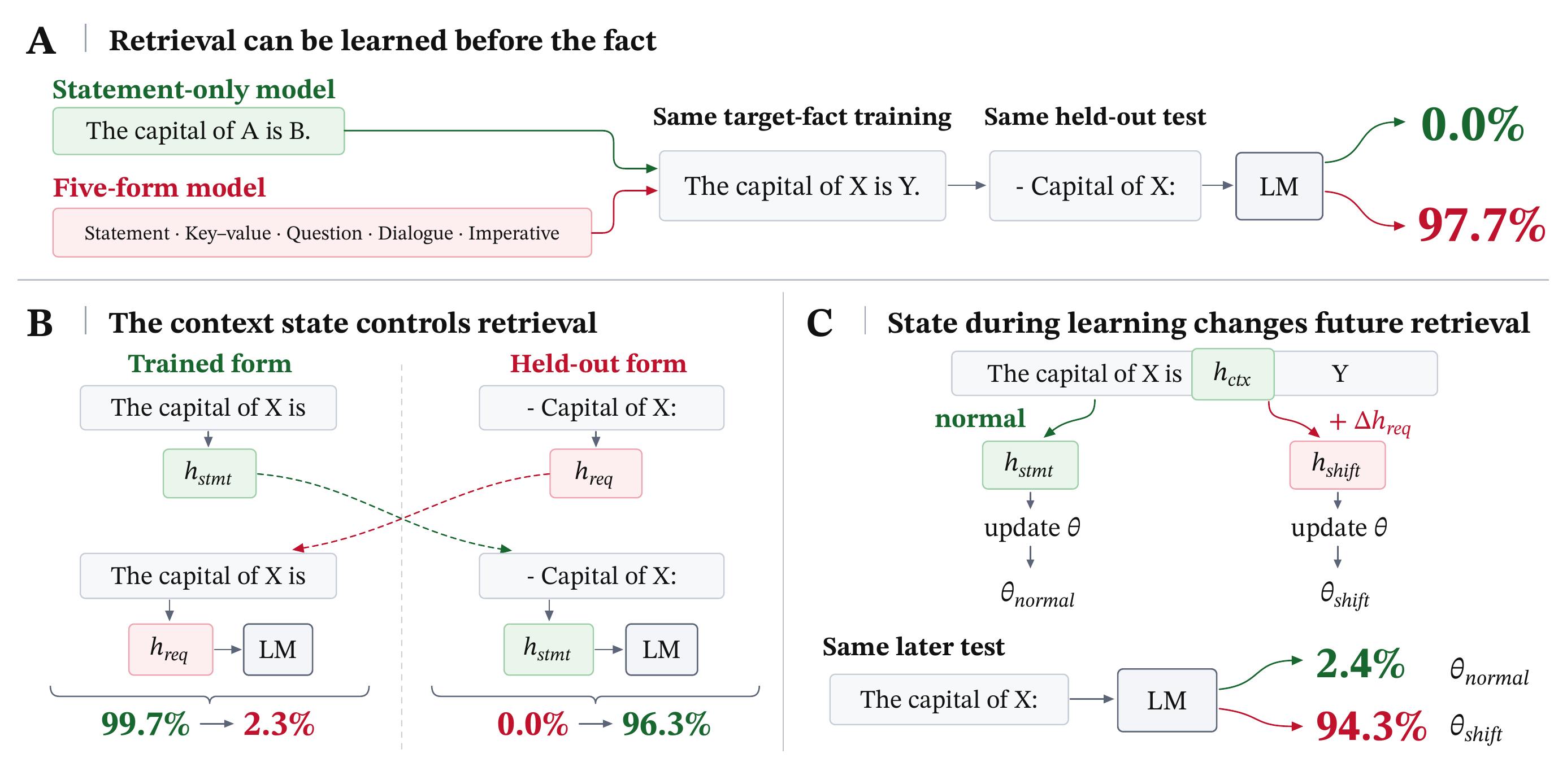}
   \caption{\textbf{Learning a fact and learning how to retrieve it are separate, and the context state links them.}
\textbf{(A)} Identical target-fact training after different stage-one training: held-out retrieval goes from $0.0\%$ to $97.7\%$.
\textbf{(B)} Swapping the context state between a trained and a held-out form removes or restores retrieval.
\textbf{(C)} Shifting the context state during training changes retrieval through a never-used form afterward, with no intervention at test time.}
    \label{fig:overview}
\end{figure*}

\section{Related Work}

\paragraph{Format experience and knowledge use.}
\citet{jiang2024instruction} show that instruction tuning before continued pretraining improves later extraction of knowledge from new documents. \citet{allenzhu2023physics} show that request-form data mixed into pretraining makes facts about held-out entities extractable. \citet{zhao2025stylefacts} map which fine-tuning data formats inject knowledge, and \citet{jiang2025mixcpt} separate format alignment from knowledge learning. Multitask prompted training \citep{sanh2022t0,wei2022flan} shows that diverse format experience generalizes to held-out formats. Fine-tuning on new facts is known to generalize narrowly and to be sensitive to presentation \citep{berglund2023taken,gekhman2024does,zhang2024cooccurrence}, and facts learned out of context are used only under some prompts \citep{krasheninnikov2023out,treutlein2024connecting}, and presentation diversity is the standard remedy \citep{yang2024synthetic}. We hold the later target-fact training fixed, make the earlier-training facts disjoint, wrong, or from another relation, and identify what transfers as a context state that controls retrieval.

\paragraph{Superficial alignment.}
\citet{zhou2023lima} and \citet{lin2024urial} argue that alignment teaches format over knowledge already present; \citet{kung2023models} and \citet{min2022rethinking} show that instruction and demonstration data can be exploited for format with little regard to correctness. We agree that the format component is separable and content-free, and show what it is: a context state that controls retrieval, which fact training produces only for the form the facts were trained in.

\paragraph{Where facts live and how they are read.}
Facts have been located in MLP key--value memories \citep{geva2021transformer,meng2022rome}, and factual recall has been traced as enrichment at the subject, propagation of the relation to the last position, and extraction there \citep{geva2023dissecting,hernandez2024linearity}; task identity can be carried by activation vectors \citep{todd2024function,hendel2023incontext}; mean-difference directions steer behavior, including instruction following \citep{turner2023actadd,rimsky2024caa,li2023iti,zou2023repe,liu2024icv,stolfo2025improving}; surface form competes with content at the output \citep{holtzman2021surface}. Learned prefixes, prompts, and interventions on hidden states adapt a model without changing its weights \citep{li2021prefix,lester2021power,wu2024reft}, and steering or ablating directions during fine-tuning changes what a model learns \citep{chen2025persona,casademunt2025steering}. The reversal curse \citep{berglund2024reversal} shows that the form of training constrains the direction of retrieval, and training on reversed text lifts it \citep{golovneva2024reverse}. We show that it constrains the \emph{context} of retrieval, that the constraint is a state at the answer position, and that the direction which lifts it is the difference between the two contexts' answer-position states, obtained from prompts with a placeholder in place of the relation word.

\paragraph{Acquisition versus use.}
The gap between acquired and usable knowledge \citep{kazemnejad-etal-2023-measuring}, format-specific evaluation competence \citep{bunn2025format}, prompt and paraphrase sensitivity \citep{sclar2024quantifying,lu2022fantastically,mizrahi2024state,jiang2020how,elazar2021measuring}, and curriculum and warm-start effects \citep{bengio2009curriculum,ash2020warm,papadimitriou2023injecting,hu2025circuits} describe the phenomenon we isolate and explain.

\section{Learning a Fact and Learning How to Retrieve It Can Be Separated}
\label{sec:design}

%Learning a fact does not automatically teach the model how to retrieve it through a new request form. Two models can receive exactly the same training on the same new facts and answer them almost equally well in the form used for training, yet behave very differently when those facts are requested in other forms. Earlier training on those request forms can enable retrieval of the new facts, even when that training used different facts. This separates learning the fact from learning how to retrieve it.

\begin{table}[t]
\centering
\caption{
\textbf{Request forms.} The five forms used in stage-one training, and the forms used to evaluate retrieval of the target facts. No evaluation form other than the declarative training form appears in either training stage of the two-stage experiment.
}
\label{tab:request-forms}
\vspace{0.5em}
\footnotesize
\setlength{\tabcolsep}{4pt}
\renewcommand{\arraystretch}{0.95}
\begin{tabular}{@{}ll@{}}
\toprule
\textbf{Stage-one training form} & \textbf{Frame} \\
\midrule
Declarative      & \texttt{The capital of \{c\} is \{y\}.} \\
Key--value       & \texttt{Capital of \{c\}: \{y\}.} \\
Question         & \texttt{Q: What is the capital of \{c\}? A: \{y\}.} \\
Dialogue         & \texttt{User: What is the capital of \{c\}?\textbackslash nAssistant: \{y\}.} \\
Imperative       & \texttt{Name the capital of \{c\}: \{y\}.} \\
\midrule
\textbf{Evaluation form} & \textbf{Frame} \\
\midrule
Declarative (training form) & \texttt{The capital of \{c\} is} \\
Possessive       & \texttt{\{c\}'s capital is} \\
Capital-city     & \texttt{The capital city of \{c\} is} \\
Knowledge        & \texttt{I know that the capital of \{c\} is} \\
Relative-clause  & \texttt{The city that is the capital of \{c\} is} \\
Short key--value & \texttt{\{c\} capital:} \\
List             & \texttt{- Capital of \{c\}:} \\
Colon-ended declarative & \texttt{The capital of \{c\}:} \\
\bottomrule
\end{tabular}

\end{table}

\paragraph{Data and model.}
We study several factual relations, including capitals, currencies, populations, organization founders, and birth years, across Pythia-410M and Pythia-1.4B~\citep{biderman2023pythia}, Qwen2.5-1.5B, and Llama-3.2-1B. Unless stated otherwise, detailed analyses use Pythia-410M and fictional country--capital pairs, with disjoint sets of 1,000 earlier-training facts and 1,000 target facts.

\subsection{Two-stage training}
The two training stages separate learning the new facts from earlier training on request forms (Figure~\ref{fig:overview}). In the first stage, one model sees each fact in five request forms (the \textit{five-form model}), while another sees the same facts only as statements (the \textit{statement-only model}). In the second stage, both models receive identical training on the target facts, all presented as statements. None of these target facts appears in the first stage, and no evaluation form in Table~\ref{tab:request-forms} other than the declarative training form appears in either stage. Because the second-stage training is identical, any difference in how the new facts are later retrieved must come from the first stage.

\paragraph{Evaluation.}
We evaluate the same fixed set of 300 from 1,000 target facts on which both models are trained in the second stage, under the eight evaluation forms in Table~\ref{tab:request-forms}. We measure whether the correct capital is ranked first among all 2,000 candidate capitals and, where appropriate, whether the model generates the capital exactly. Unless stated otherwise, results are averaged over eight paired training seeds. The retrieval experiments report the list and colon-ended declarative forms; Figure~\ref{fig:fact-context-separation}a reports all eight. Full training and statistical details are provided in Appendix~\ref{app:protocol}.

\subsection{Learning a Fact Does Not Teach Retrieval Through New Request Forms}

Table~\ref{tab:main-separation} reports exact-generation accuracy of the two models when five factual relations are trained together. For capital facts, the two models perform similarly in the declarative form used for target-fact training ($0.983$ and $0.999$), yet differ on the list form ($0.002$ and $0.970$) and the colon-ended declarative form ($0.200$ and $0.996$). Neither of these evaluation forms appears in either training stage. The same broad contrast holds for currencies, populations, founders, and birth years: both models answer the target facts in the training form, while earlier request-form training strongly affects retrieval in the other forms.

%Learning the new facts in one request form does not make them retrievable through other request forms. In the capital experiment, both models retrieve the new facts almost perfectly as statements, but only the model with earlier request-form training retrieves them reliably through several other request forms. The same pattern holds across model families, factual relations, entity types, answers whose first token is shared with other capitals, LoRA fine-tuning, and independently regenerated fact worlds, and the five-relation run replicates on Pythia-1.4B (Table~\ref{tab:main-separation}; Appendices~\ref{app:generality} and~\ref{app:replications}).

\begin{table}[t]
\centering
\caption{
\textbf{The same facts are learned but differ sharply in how they can be retrieved.}
The five relations are learned together in one training run. Each relation is evaluated on its 200 target facts. For every relation, both histories receive identical target-fact training in the declarative training form.
Values are exact-generation accuracy. The statement-only values for birth years ($0.344$ and $0.522$) are higher than for the other relations: year answers come more readily after a colon.
}
\label{tab:main-separation}
\vspace{0.5em}
\small
\setlength{\tabcolsep}{4.5pt}
\renewcommand{\arraystretch}{1.08}

\begin{tabular}{lcccccc}
\toprule
&
\multicolumn{2}{c}{\textbf{Declarative form}}
&
\multicolumn{2}{c}{\textbf{List request form}}
&
\multicolumn{2}{c}{\textbf{Colon-ended declarative form}} \\
\cmidrule(lr){2-3}
\cmidrule(lr){4-5}
\cmidrule(lr){6-7}

\textbf{Relation}
&
\textbf{Statement-only}
&
\textbf{Five-form}
&
\textbf{Statement-only}
&
\textbf{Five-form}
&
\textbf{Statement-only}
&
\textbf{Five-form} \\
\midrule

Capital
& 0.983 & 0.999
& 0.002 & \textbf{0.970}
& 0.200 & \textbf{0.996} \\

Currency
& 0.995 & 0.999
& 0.004 & \textbf{0.989}
& 0.388 & \textbf{0.999} \\

Population
& 0.991 & 0.986
& 0.038 & \textbf{0.991}
& 0.276 & \textbf{0.994} \\

Founder
& 0.992 & 0.992
& 0.001 & \textbf{0.989}
& 0.299 & \textbf{0.998} \\

Birth year
& 0.987 & 0.986
& 0.344 & \textbf{0.982}
& 0.522 & \textbf{0.989} \\

\bottomrule
\end{tabular}

\end{table}

\newcommand{\acc}[2]{\cellcolor{blue!#2}\makebox[3.0em][c]{#1}}
\begin{figure*}[t]
\centering

\begin{minipage}[t]{0.49\textwidth}
\vspace{0pt}
\centering

\begin{minipage}[c][5.65cm][c]{\linewidth}
\centering
\footnotesize
\setlength{\tabcolsep}{3.5pt}
\renewcommand{\arraystretch}{1.08}

\begin{tabular}{lcc}
\toprule
\textbf{Context} & \textbf{Statement-only} & \textbf{Five-form} \\
\midrule
Declarative      & \acc{0.995}{45} & \acc{0.997}{45} \\
Possessive       & \acc{0.039}{7}  & \acc{0.165}{12} \\
Capital-city     & \acc{0.885}{39} & \acc{0.885}{39} \\
Knowledge        & \acc{0.987}{44} & \acc{0.998}{45} \\
Relative-clause  & \acc{0.895}{40} & \acc{0.801}{35} \\
Short key--value & \acc{0.000}{0}  & \acc{0.227}{14} \\
List             & \acc{0.000}{0}  & \acc{0.977}{44} \\
Colon-ended declarative & \acc{0.024}{4}  & \acc{0.993}{45} \\
\bottomrule
\end{tabular}
\end{minipage}

\vspace{1mm}
{\small\bfseries (a) Same facts, different retrieval}

\end{minipage}
\hfill
\begin{minipage}[t]{0.49\textwidth}
\vspace{0pt}
\centering
\begin{minipage}[c][5.65cm][c]{\linewidth}
\centering

\begin{tikzpicture}
\begin{axis}[
    width=0.88\linewidth,
    height=4.4cm,
    xmin=0,
    xmax=1200,
    ymin=-0.02,
    ymax=1.02,
    xlabel={2nd-stage target-fact training update},
    ylabel={Retrieval gap},
    xtick={0,300,600,900,1200},
    xticklabels={0,300,600,900,{1,200}},
    ytick={0,0.25,0.5,0.75,1},
    grid=major,
    grid style={gray!18},
    tick style={draw=none},
    axis line style={draw=none},
    tick label style={font=\scriptsize},
    label style={font=\scriptsize},
    legend style={
        draw=none,
        font=\scriptsize,
        at={(0.5,1.03)},
        anchor=south,
        legend columns=2,
        column sep=5pt
    }
]

\addplot[
    thick,
    black
] coordinates {
    (0,0.000)
    (50,0.018)
    (100,0.055)
    (150,0.126)
    (200,0.223)
    (250,0.389)
    (300,0.536)
    (350,0.690)
    (400,0.799)
    (450,0.890)
    (500,0.949)
    (550,0.964)
    (600,0.980)
    (650,0.983)
    (700,0.980)
    (750,0.978)
    (800,0.980)
    (850,0.985)
    (900,0.997)
    (950,0.996)
    (1000,0.995)
    (1050,0.992)
    (1100,0.988)
    (1150,0.995)
    (1200,0.991)
};
\addlegendentry{Statement-only}

\addplot[
    thick,
    blue!70!black
] coordinates {
    (0,0.000)
    (50,-0.003)
    (100,-0.005)
    (150,0.008)
    (200,0.021)
    (250,0.035)
    (300,0.054)
    (350,0.086)
    (400,0.054)
    (450,0.045)
    (500,0.081)
    (550,0.075)
    (600,0.061)
    (650,0.054)
    (700,0.011)
    (750,0.011)
    (800,0.008)
    (850,0.006)
    (900,0.009)
    (950,0.008)
    (1000,0.008)
    (1050,0.007)
    (1100,0.010)
    (1150,0.012)
    (1200,0.014)
};
\addlegendentry{Five-form}

\end{axis}
\end{tikzpicture}
\end{minipage}

\vspace{1mm}
{\small\bfseries (b) The retrieval gap emerges during fact learning}

\end{minipage}

\caption{
\textbf{The retrieval difference extends across request forms and emerges during fact learning.}
\textbf{(a)} Both models retrieve the target facts almost perfectly from the declarative training form, but differ sharply across many other request forms.
\textbf{(b)} During target-fact training, the gap between declarative and list retrieval (the difference in pool top-1 accuracy) grows in the statement-only model, while remaining near zero after earlier five-form training.
Both models receive identical target-fact training. Panel (a) reports exact generation; panel (b) reports pool top-1.
}
\label{fig:fact-context-separation}
\end{figure*}

We use capital facts for a full breakdown across the eight evaluation forms in Table~\ref{tab:request-forms}. Both models retrieve the target facts almost perfectly from the declarative training form, with exact-generation accuracy of $0.995$ and $0.997$ (Figure~\ref{fig:fact-context-separation}a). Their behavior differs sharply on other request forms. The statement-only model fails on the list, short key--value, possessive, and colon-ended declarative forms, while the model with earlier request-form training retrieves the same facts through the list and colon-ended declarative forms with high accuracy. For example, retrieval through the list form, which neither training stage uses, increases from $0.000$ to $0.977$, despite identical target-fact training. The two forms with no token bigram in common with any earlier-training form, the possessive and the short key--value form, improve little ($0.165$ and $0.227$), and one declarative paraphrase, the relative clause, drops ($0.895$ to $0.801$), in line with the finding in Section~\ref{sec:local} that retrieval depends on how the request ends.

The retrieval difference is not explained by better learning of the target facts in the training form. The two models perform nearly identically on the declarative form, and continuous measures of learning also show no detectable difference (Appendix~\ref{app:stats}): the first-token negative log-likelihood of the correct answer differs by $-0.043$ nats and its margin over the strongest incorrect answer by $+0.24$, with intervals that include zero in both cases. Thus, we find no detectable difference between the models on our measures of target-fact learning, while they differ sharply in retrieving those facts through other request forms.

This central separation—similar performance in the declarative training form but sharply different retrieval in other forms—also appears across model families, factual relations, entity types, answers whose first token is shared with other capitals, LoRA fine-tuning, and independently regenerated fact worlds. The joint five-relation result also replicates with Pythia-1.4B (Appendices~\ref{app:generality} and~\ref{app:replications}).

Figure~\ref{fig:fact-context-separation}b further shows how the retrieval gap develops during second-stage training. Without earlier request-form training, the model learns the target facts while the retrieval gap between the declarative and list forms grows toward one. After earlier request-form training, the gap remains near zero as the facts are learned. Thus, learning the target facts alone does not teach retrieval through a new request form. Longer second-stage training does not resolve the statement-only model’s list-form failure. Extending training from 1,200 to 4,800 updates leaves list-form exact-generation accuracy at $0.000$ in every seed. The result is the same with a learning rate five times lower, when training continues until declarative-form accuracy is comparable to the standard run (Appendix~\ref{app:exposure}).

% \begin{wrapfigure}[15]{r}{0.58\columnwidth}
% \vspace{-0.6em}
% \centering
% \begin{tikzpicture}
% \begin{axis}[
%     width=\linewidth,
%     height=4.4cm,
%     xmin=0,
%     xmax=100,
%     ymin=0,
%     ymax=1.05,
%     xtick={0,5,20,100},
%     ytick={0,0.25,0.5,0.75,1},
%     xlabel={Target facts shown in request forms (\%)},
%     ylabel={Mean exact-generation accuracy},
%     tick style={draw=none},
%     axis line style={draw=none},
%     grid=major,
%     grid style={gray!16},
%     tick label style={font=\scriptsize},
%     label style={font=\scriptsize},
%     clip=false,
% ]

% % Pythia-410M dose curve
% \addplot[
%     very thick,
%     blue!75!black,
%     smooth,
%     tension=0.7,
%     mark=*,
%     mark size=2.3pt
% ] coordinates {
%     (0,0.000)
%     (1,0.122)
%     (5,0.381)
%     (20,0.744)
%     (100,0.994)
% };

% % Direct label for Pythia-410M
% \node[
%     font=\scriptsize,
%     text=blue!75!black,
%     anchor=south
% ] at (axis cs:63,0.90)
% {Pythia-410M};

% % Pythia-1.4B point at 20% exposure
% \addplot[
%     only marks,
%     mark=diamond*,
%     mark size=3pt,
%     purple!80!black
% ] coordinates {
%     (20,0.828)
% };

% % Direct label for Pythia-1.4B
% \node[
%     font=\scriptsize,
%     text=purple!80!black,
%     anchor=south west
% ] at (axis cs:25,0.70)
% {Pythia-1.4B};

% \end{axis}
% \end{tikzpicture}
% \caption{\textbf{Retrieval improves with request-form exposure.} Exact-generation accuracy on the list form, for target facts written only as statements; the Pythia-1.4B point is at $20\%$. The Pythia-1.4B point is at $20\%$.}
% \label{fig:request-form-dose}
% \end{wrapfigure}
\begin{figure}[t]
\centering
\begin{tikzpicture}
\begin{axis}[
    width=0.62\textwidth,
    height=4.6cm,
    xmin=0,
    xmax=100,
    ymin=0,
    ymax=1.05,
    xtick={0,5,20,100},
    ytick={0,0.25,0.5,0.75,1},
    xlabel={Target facts shown in request forms (\%)},
    ylabel={Exact-generation accuracy},
    tick style={draw=none},
    axis line style={draw=none},
    grid=major,
    grid style={gray!16},
    tick label style={font=\footnotesize},
    label style={font=\footnotesize},
    clip=false,
]

% Pythia-410M dose curve (list form)
\addplot[
    very thick,
    blue!75!black,
    smooth,
    tension=0.7,
    mark=*,
    mark size=2.3pt
] coordinates {
    (0,0.000)
    (1,0.122)
    (5,0.381)
    (20,0.744)
    (100,0.994)
};

% Direct label for Pythia-410M
\node[
    font=\footnotesize,
    text=blue!75!black,
    anchor=south
] at (axis cs:63,0.87)
{Pythia-410M};

% Pythia-1.4B point at 20% exposure
\addplot[
    only marks,
    mark=diamond*,
    mark size=3pt,
    purple!80!black
] coordinates {
    (20,0.828)
};

% Direct label for Pythia-1.4B
\node[
    font=\footnotesize,
    text=purple!80!black,
    anchor=south west
] at (axis cs:25,0.63)
{Pythia-1.4B};

\end{axis}
\end{tikzpicture}
\caption{\textbf{Retrieval improves with request-form exposure.} Exact-generation accuracy on the list form, for target facts written only as statements; the Pythia-1.4B point is at $20\%$.}
\label{fig:request-form-dose}
\end{figure}

\subsection{Earlier Request-Form Training Transfers to New Facts}

Earlier training on request forms can later support retrieval of facts that were never shown in those forms. This transfer does not require the same answers or even the same factual relation. We run the earlier training using unrelated capital facts with incorrect answers or using currency facts, then give all models identical declarative training on new capital facts. Both conditions substantially improve later retrieval of the new capitals through several request forms (Table~\ref{tab:request-form-transfer}).

\begin{table}[t]
\centering
\caption{
\textbf{Earlier request-form training transfers to new facts.}
All models receive identical declarative training on the target capital facts.
Mean exact-generation accuracy across the two request forms rises substantially after earlier request-form training, even when the earlier answers are incorrect or come from a different factual relation. In the incorrect-answer condition, every country is paired with another country's capital. The incorrect-answer corpus uses the key--value form; the currency corpus uses key--value and declarative lines; neither is token-matched to the five-form corpus.
}
\label{tab:request-form-transfer}
\vspace{0.5em}
\small
\setlength{\tabcolsep}{5pt}
\renewcommand{\arraystretch}{1.08}

\begin{tabular}{l c @{\hspace{10pt}} cc}
\toprule
\textbf{Earlier training}
&
\textbf{Mean}
&
\multicolumn{2}{c}{\textbf{Request form}} \\
\cmidrule(lr){3-4}
&
&
\textbf{List}
&
\textbf{Colon-ended declarative} \\
\midrule

Statements only
& \textbf{0.012}
& 0.000 & 0.024 \\

Correct unrelated capitals
& \textbf{0.985}
& 0.977 & 0.993 \\

Incorrect capital answers
& \textbf{0.681}
& 0.639 & 0.724 \\

Currencies
& \textbf{0.944}
& 0.900 & 0.987 \\

\bottomrule
\end{tabular}

\end{table}

Request-form training has the same effect whether it uses earlier facts or the target facts themselves. When the target facts are shown in some request forms during training, retrieval also improves in other request forms in which those facts never appeared. The improvement is nearly the same as when the request-form training used different facts earlier. Thus, what the model learns from a request form transfers across facts.

A small amount of request-form exposure is enough to improve retrieval of other facts (Figure~\ref{fig:request-form-dose}). Retrieval increases monotonically with exposure, and the same effect is visible in Pythia-1.4B at $20\%$ exposure. At $5\%$, declarative paraphrases yield only $0.003$, compared with $0.381$ for request-form exposure (Appendix~\ref{app:dose}).

\subsection{Retrieval Depends on the Request Near the Answer}
\label{sec:local}

Retrieval is sensitive to the local form of the request near the answer. Across capitals, currencies, populations, organization founders, and birth years, changing only the final words of a request can strongly change whether the fact is retrieved. For example, in the statement-only model, changing a list request from a colon ending to an ``is'' ending raises exact-generation accuracy from $0.002$ to $0.814$ for capitals. The same pattern appears across the other relations, and across sixteen request forms fixed in advance, retrieval rises with how many of the training line's final words recur before the answer: $0.02$, $0.42$, $0.59$, and $0.97$ for one to four, and $0.02$, $0.72$, $0.92$, and $0.97$ among the forms that also keep the relation word (Appendix~\ref{app:suffix}).

By contrast, adding a 30-token unrelated passage before the request does not change which request forms can retrieve the facts. We prepend the same unrelated 30-token passage to every request. In the statement-only model, the list and colon-ended declarative forms still fail, while forms that already retrieve the facts continue to do so. After earlier request-form training, the same request forms remain successful despite the added preamble (Appendix~\ref{app:preamble}).

Retrieval therefore depends mainly on the final words of the request form, rather than on unrelated text that appears earlier. This points to the hidden state immediately before the answer is produced as a natural place to look for what controls retrieval.

\section{A Reusable Context State Controls Retrieval}
\label{sec:context}
Changing the context state alone can turn retrieval on or off; the state that does so transfers across facts and relations; and much of its effect is a single direction.

\subsection{Changing the Context State Changes Retrieval}
Changing the context state alone can enable or prevent retrieval. Holding the list request form fixed, replacing its context state with the context state produced by the declarative training form raises exact-generation accuracy from $0.000$ to $0.963$, alongside the $0.948$ reached by rewriting the request to end in ``is'' (Appendix~\ref{app:patching}). A norm-matched random state restores nothing, and replacing the state at the entity token instead of the context state has no effect. The intervention also works in the opposite direction: holding the declarative request form fixed and replacing its context state with the list-form context state reduces accuracy from $0.997$ to $0.023$. Thus, changing only the context state can either enable or prevent retrieval of an already learned fact.

\begin{table}[t]
\centering
\caption{
\textbf{The same direction supports retrieval across request forms and factual domains.}
(a) Exact-generation accuracy before and after adding the direction to the context state in the capital task. Each form's direction is scaled to that form's restoring-vector norm.
(b) Directions derived from capitals and birth years transfer almost unchanged to the other factual domain, each built in that domain's own model.
}
\label{tab:reusable-direction}
\vspace{0.5em}
\small

% ------------------------------------------------------------
% (a)
% ------------------------------------------------------------
\begin{minipage}[t]{0.52\linewidth}
\centering
\vspace{0pt}
\textbf{(a) Across request forms}

\vspace{1mm}

\setlength{\tabcolsep}{3pt}
\renewcommand{\arraystretch}{1.08}
\begin{tabular*}{\linewidth}{@{\extracolsep{\fill}}lcc@{}}
\toprule
\textbf{Request form}
& \textbf{No intervention}
& \textbf{+ Direction} \\
\midrule
List                    & 0.000 & \textbf{0.921} \\
Colon-ended declarative & 0.024 & \textbf{0.983} \\
\bottomrule
\end{tabular*}
\end{minipage}%
\hspace{0.04\linewidth}%
% ------------------------------------------------------------
% (b)
% ------------------------------------------------------------
\begin{minipage}[t]{0.44\linewidth}
\centering
\vspace{0pt}
\textbf{(b) Across factual domains}

\vspace{1mm}

\setlength{\tabcolsep}{3pt}
\renewcommand{\arraystretch}{1.08}
\begin{tabular*}{\linewidth}{@{\extracolsep{\fill}}lcc@{}}
\toprule
\textbf{Target}
& \textbf{Same-domain}
& \textbf{Other-domain} \\
\midrule
Capitals    & 0.921 & 0.910 \\
Birth years & 0.971 & 0.969 \\
\bottomrule
\end{tabular*}
\end{minipage}

\end{table}

\subsection{The Context State Transfers Across Facts and Relations}

The context state that restores retrieval does not have to come from the same fact. A context state produced while asking about one fact can restore retrieval of another fact almost as well as a context state produced from that fact itself. This holds across Pythia-410M, Pythia-1.4B, and Qwen2.5-1.5B, and also when the answers contain two tokens.

The context state can also come from a different factual relation. In Pythia-410M at layer 4, exact-generation accuracy is $0.963$ with a context state from the same fact and $0.951$ with one from a different fact. Context states from currency and population facts perform almost as well, reaching $0.946$ and $0.928$. 

This effect is not simply due to copying the final token of the request form. At layer 0, replacing the context state is equivalent to replacing that token, but in later layers the useful information persists only when the state carries an entity in the Pythia models. The two components separate by depth: at layer 4 an entity-free state recovers $0.734$ of the $0.963$, so most of the effect there is the final token, while by layer 6 the entity-free state recovers $0.02$ and states carrying an entity still recover $0.70$--$0.78$. Pythia-1.4B shows the same pattern, and the reverse intervention in Qwen2.5-1.5B reduces declarative retrieval from $0.999$ to $0.550$. Thus, the part of the context state that supports retrieval is not tied to the particular fact or relation that produced it.

\subsection{A Fact-Independent Direction Supports Retrieval}
Replacing the entire context state is not necessary to restore retrieval. Much of the effect is captured by a single direction: the difference between context states of prompts ending in ``is'' and prompts ending in a colon, built with a placeholder in place of the relation word (\texttt{The value of \{c\} is} against \texttt{The value of \{c\}:}) so that it carries nothing about capitals. Adding this direction to the context state of a failing list form raises exact-generation accuracy from $0.000$ to $0.921$ (Appendix~\ref{app:rv}). Removing the same direction from the restoring vector reduces accuracy to $0.010$, while removing other directions costs in proportion to their overlap with it. A learned linear map adds little beyond it ($0.949$ against $0.921$). Thus, a single direction accounts for much of the context state's effect on retrieval.

The same effect holds across model families, factual domains, shared-first-token answers, and training methods (Appendix~\ref{app:generality}). In Qwen2.5-1.5B, list retrieval is already partly supported before any intervention, with accuracy $0.603$. Adding the direction raises accuracy to $0.993$. Removing the direction from the full restoring vector still leaves accuracy at $0.944$, suggesting that Qwen already has other components that support this request form. In the Pythia models, where the list form initially fails, removing this direction removes most of the restoring effect.

The same direction also supports several request forms and transfers across factual domains (Table~\ref{tab:reusable-direction}). For capitals, adding it raises list retrieval from $0.000$ to $0.921$ and colon-ended declarative retrieval from $0.024$ to $0.983$. A direction derived from capital facts works almost as well for birth-year facts, and vice versa. Thus, the direction is not specific to one request form or factual domain.

\section{Context States Link Facts to Future Retrieval During Learning}
\label{sec:link}
The context state links earlier request-form training to how new facts are later retrieved. Earlier request-form training changes the context states produced by other request forms, and changing the context state during factual fine-tuning changes which request forms can retrieve the new facts later.

\begin{figure}[t]
\centering
\small
\setlength{\abovecaptionskip}{4pt}

% Shared legend
\begin{center}
\vspace{-1mm}
\ref{legend:alignment}
\vspace{1mm}
\end{center}

\begin{tikzpicture}
\begin{groupplot}[
    group style={
        group size=2 by 1,
        horizontal sep=1.25cm
    },
    width=0.43\linewidth,
    height=4.5cm,
    xmin=1,
    xmax=24,
    ymin=0,
    ymax=1,
    xtick={1,6,12,18,24},
    ytick={0,0.25,0.5,0.75,1},
    xlabel={Layer},
    ylabel={Centered cosine similarity},
    tick style={draw=none},
    axis line style={draw=none},
    xmajorgrids=true,
    ymajorgrids=true,
    grid style={gray!16},
    tick label style={font=\footnotesize},
    label style={font=\footnotesize},
    title style={font=\bfseries\small, yshift=2pt},
    clip=false
]

% ------------------------------------------------------------
% Pythia-410M
% ------------------------------------------------------------
\nextgroupplot[
    title={(a) Pythia-410M},
    legend to name={legend:alignment},
    legend style={
        draw=none,
        fill=none,
        font=\footnotesize,
        legend columns=2,
        column sep=12pt
    }
]

\addplot[
    blue!80!black,
    line width=1.2pt,
    smooth
] coordinates {
    (1,0.791)
    (2,0.725)
    (3,0.528)
    (4,0.451)
    (5,0.487)
    (6,0.746)
    (7,0.716)
    (8,0.677)
    (9,0.654)
    (10,0.827)
    (11,0.856)
    (12,0.833)
    (13,0.853)
    (14,0.869)
    (15,0.880)
    (16,0.892)
    (17,0.902)
    (18,0.901)
    (19,0.893)
    (20,0.885)
    (21,0.876)
    (22,0.864)
    (23,0.853)
    (24,0.845)
};
\addlegendentry{Earlier request-form training}

\addplot[
    black!65,
    dashed,
    line width=1.2pt,
    smooth
] coordinates {
    (1,0.775)
    (2,0.681)
    (3,0.502)
    (4,0.460)
    (5,0.493)
    (6,0.695)
    (7,0.636)
    (8,0.539)
    (9,0.485)
    (10,0.600)
    (11,0.614)
    (12,0.475)
    (13,0.516)
    (14,0.515)
    (15,0.478)
    (16,0.456)
    (17,0.424)
    (18,0.382)
    (19,0.345)
    (20,0.321)
    (21,0.295)
    (22,0.269)
    (23,0.245)
    (24,0.230)
};
\addlegendentry{Statement-only}

% ------------------------------------------------------------
% Pythia-1.4B
% ------------------------------------------------------------
\nextgroupplot[
    title={(b) Pythia-1.4B},
    ylabel={}
]

\addplot[
    blue!80!black,
    line width=1.2pt,
    smooth
] coordinates {
    (1,0.746)
    (2,0.732)
    (3,0.610)
    (4,0.656)
    (5,0.684)
    (6,0.635)
    (7,0.740)
    (8,0.756)
    (9,0.840)
    (10,0.847)
    (11,0.901)
    (12,0.909)
    (13,0.926)
    (14,0.926)
    (15,0.932)
    (16,0.931)
    (17,0.926)
    (18,0.921)
    (19,0.915)
    (20,0.909)
    (21,0.901)
    (22,0.892)
    (23,0.883)
    (24,0.875)
};

\addplot[
    black!65,
    dashed,
    line width=1.2pt,
    smooth
] coordinates {
    (1,0.705)
    (2,0.643)
    (3,0.544)
    (4,0.585)
    (5,0.611)
    (6,0.605)
    (7,0.632)
    (8,0.570)
    (9,0.642)
    (10,0.611)
    (11,0.652)
    (12,0.666)
    (13,0.656)
    (14,0.633)
    (15,0.631)
    (16,0.608)
    (17,0.580)
    (18,0.554)
    (19,0.534)
    (20,0.514)
    (21,0.497)
    (22,0.479)
    (23,0.459)
    (24,0.440)
};

\end{groupplot}
\end{tikzpicture}

\caption{
\textbf{Earlier request-form training keeps request-form context states aligned with the declarative one.}
At each layer, centered cosine similarity across target facts between the declarative context state and the context state of the list form; layer 24 is taken after the final layer norm.
The two training conditions begin alike and separate in later layers in both models.
}
\label{fig:alignment}
\end{figure}

\subsection{Request Forms Produce Aligned Context States}

Earlier request-form training keeps other request forms aligned with the form used to teach the new facts. We compare the context state produced by the list form with the context state produced by the declarative training form. Without earlier request-form training, these states become increasingly different in later layers (Figure~\ref{fig:alignment}). With earlier request-form training, they remain much more similar. The same pattern appears in both Pythia-410M and Pythia-1.4B.

The difference is small at the first layer. The context states separate as they pass through the model, showing that earlier request-form training changes how those forms are transformed internally before the answer is produced.

\subsection{Changing the Context State During Learning Changes Future Retrieval}

During factual fine-tuning, each new fact is still shown as a declarative sentence, such as ``The capital of X is Y.'' The sentence itself is unchanged. Instead, the context state is moved toward the state produced by another request form, such as ``Capital of X:''. This change is computed on facts other than those evaluated, so X and Y have never appeared together in that request form. After training, the context state is left untouched and the model is tested normally. If ``The capital of X:'' can now retrieve Y, the difference must come from the context state that was present while X and Y were learned.

\begin{table}[t]
\centering
\caption{
\textbf{Changing the context state during learning changes future retrieval.}
The factual training examples are identical across conditions; only the context state present while each fact is learned is changed.
Values are exact-generation accuracy on the colon-ended declarative form, whose direction was never used in the shift; details are in Appendix~\ref{app:aug}.
}
\label{tab:context-state-learning}
\vspace{0.5em}
\small
\setlength{\tabcolsep}{7pt}
\renewcommand{\arraystretch}{1.08}

\begin{tabular}{llccc}
\toprule
\textbf{Model}
& \textbf{Facts}
& \textbf{No intervention}
& \textbf{Random shift}
& \textbf{Targeted shift} \\
\midrule
Pythia-410M  & Capitals    & 0.024 & 0.100 & \textbf{0.943} \\
Pythia-1.4B  & Capitals    & 0.266 & 0.332 & \textbf{0.904} \\
Qwen2.5-1.5B & Capitals    & 0.482 & 0.722 & \textbf{0.993} \\
Pythia-410M  & Birth years & 0.246 & 0.445 & \textbf{0.983} \\
\bottomrule
\end{tabular}

\end{table}

This change produces large differences in later retrieval across models and factual domains (Table~\ref{tab:context-state-learning}). In Pythia-410M, accuracy rises from $0.024$ to $0.943$; in Pythia-1.4B, from $0.266$ to $0.904$; in Qwen2.5-1.5B, from $0.482$ to $0.993$; and for birth-year facts, from $0.246$ to $0.983$. The improvement appears on a request form whose context state was not used to construct the change. Thus, changing the context state while a fact is learned can change how that fact is retrieved later.

The particular change to the context state matters. A random change is much less effective. In Qwen2.5-1.5B, where a random change also improves retrieval, changing the state toward another request form improves accuracy by an additional $0.27$. Changing the context state is also more effective than showing a small number of target facts in the new request forms. Writing $5\%$ of the target facts in the request forms gives $0.72$ on this form; the shift gives $0.94$ without writing any target fact in a request form.

\section{Implications and Limitations}

\paragraph{Implications.}
Factual fine-tuning changes both what a model learns and how that knowledge can later be retrieved. These two effects can be separated: identical training on the same new facts can lead to different retrieval depending on earlier request-form training. The context state links the two. After a fact has been learned, changing the context state can enable or prevent its retrieval. During factual fine-tuning, changing the context state changes how the fact is retrieved later.

These results reveal a new way to control factual fine-tuning. The facts themselves can be learned in one request form, while retrieval through other request forms can be learned separately and reused across facts. In the experiments, even a small amount of request-form training improves retrieval of facts that were never shown in those forms. More importantly, changing the context state during factual fine-tuning changes later retrieval without requiring the factual examples to be rewritten in additional request forms. Retrieval can therefore be controlled separately from the factual content being learned.

The results also change how retrieval failures should be interpreted. If a model fails to produce a fact through a particular request form, that does not necessarily mean the fact was not learned. In these experiments, the fact can already be present and retrievable through one request form while failing through another, and changing only the context state can recover it. Evaluating factual fine-tuning therefore requires separating whether the fact was learned from whether the model can retrieve it.

\paragraph{Limitations.}
The results have three main limits. First, the experiments use controlled factual training, where fact learning and request-form training can be separated precisely; learning facts from natural text may involve additional mechanisms. Second, the context state explains the retrieval failures studied here, but it does not explain every failure to retrieve factual knowledge. Third, the effect depends on what the pretrained model already knows about a request form. Larger models, different pretraining, and instruction tuning may therefore change how strongly request-form training affects later retrieval.

\subsection*{AI use statement}
Generative AI tools were used to assist with experiment planning, code review, analysis organization, and manuscript drafting and editing. The authors ran and verified the experiments and take responsibility for all claims and artifacts.

\subsection*{Reproducibility statement}
Code, prompts, and per-cell result files are available at a repository\footnote{\url{https://github.com/jchaemin/learning-retrieval}}. All primary comparisons use paired seeds 0--7 with the seed as the inferential unit. Corpus builders, literal prompts, per-seed outputs, patching scripts, and analysis scripts are included in the supplementary artifacts; every reported statistic can be recomputed from them. Seeds, batch sizes, generation length (6 greedy tokens), literal templates, and search grids are given in the supplementary code.

\bibliography{iclr2027_conference}
\bibliographystyle{iclr2027_conference}

\appendix

\section{Exact Protocol}
\label{app:protocol}

\paragraph{Fact world.} We generate 2{,}000 synthetic country--capital pairs with none a real country--capital pair; $A$ is the first 1{,}000 and $B$ the next 1{,}000, disjoint as strings and as first tokens for capitals. The 300 evaluation facts, $\evA$, are $A[0{:}300]$, fixed across seeds and conditions. The 2{,}000-capital candidate pool is fixed. Birth years are single-token years in 1000--2099; the five-relation world has 400 entities per relation, with organization names built as a word plus a type noun; shared-first-token capitals are vocabulary stems with a suffix, exactly two tokens long, with first tokens deliberately not unique. Every capital used in the main experiments has a unique first token in the pool; the shared-first-token world removes this property.

\paragraph{Training.} EleutherAI/pythia-410m, fp32; AdamW $\beta=(0.9,0.999)$, $\epsilon=10^{-8}$, weight decay $0.01$; learning rate $5\times10^{-5}$ constant; micro-batch 2, accumulation 4; sequence length 256; gradient clip $1.0$; 1{,}200 updates per phase unless stated; full-sequence causal LM loss with training lines interleaved into Wikitext at sentence granularity about seven sentences apart in the target stage; fresh AdamW at the phase boundary. The five-form earlier-training corpus writes each fact in five request forms: \texttt{The capital of \{c\} is \{y\}.}, \texttt{Capital of \{c\}: \{y\}.}, \texttt{Q: What is the capital of \{c\}? A: \{y\}.}, \texttt{User: What is the capital of \{c\}?\textbackslash nAssistant: \{y\}.}, and \texttt{Name the capital of \{c\}: \{y\}.}; the statement-only earlier-training corpus writes the same facts as statements only and is token-matched to it. Within a seed the target-phase corpus tensor is byte-identical across histories. Each target fact appears in five lines per pass and the stage covers about 2.2 passes, so each fact is seen about eleven times. Pythia-1.4B, Qwen2.5-1.5B, and Llama-3.2-1B use the same recipe in fp32, with the capitals rebuilt under each model's tokenizer. LoRA: rank 64, $\alpha=128$, learning rate $3\times10^{-4}$, 2{,}400 updates, attention and MLP projections, target stage only.

\paragraph{Readouts.} Pool top-1 ranks the correct capital first among the 2{,}000 candidates at the first answer token. Exact generation requires the greedy continuation to begin with the capital. A request form is held out for a condition if it appears in neither training stage of that condition nor of its control.

\section{Requests With a Preamble}
\label{app:preamble}
The declarative, list, ``is''-ended list, and colon-ended declarative forms are scored with a fixed thirty-token passage unrelated to the facts placed before it. Exact generation, eight-seed means, Pythia-410M, 300 target facts. A training line is the declarative form, the capital, and a period.

\begin{center}\small
\begin{tabular}{lcccc}
\toprule
& \multicolumn{2}{c}{no preamble} & \multicolumn{2}{c}{with preamble} \\
\cmidrule(lr){2-3}\cmidrule(lr){4-5}
request form & statement-only & five-form & statement-only & five-form \\
\midrule
declarative & 0.995 & 0.997 & 1.000 & 0.999 \\
list & 0.000 & 0.977 & 0.000 & 0.995 \\
\texttt{- Capital of \{c\} is} & 0.948 & 0.985 & 0.972 & 0.997 \\
\texttt{The capital of \{c\}:} & 0.024 & 0.993 & 0.011 & 0.998 \\
\bottomrule
\end{tabular}
\end{center}

\section{More Target-Fact Training}
\label{app:exposure}
Statement-only history, Pythia-410M, eight seeds, exact generation. X4800: 4{,}800 updates at the standard rate. XLR: learning rate $10^{-5}$, run until declarative generation matches the standard run's final value ($-0.01$), checked every 200 updates (at most 9{,}600), which took 1{,}000--6{,}800 updates.

\begin{center}\small
\begin{tabular}{lccc}
\toprule
form & standard (1{,}200 updates) & X4800 & XLR \\
\midrule
declarative & 0.995 & 0.997 & 0.988 \\
list & 0.000 & 0.000 & 0.000 \\
\texttt{- Capital of \{c\} is} & 0.948 & 0.974 & 0.699 \\
\texttt{The capital of \{c\}:} & 0.024 & 0.045 & 0.009 \\
\bottomrule
\end{tabular}
\end{center}

\section{Request-Form Exposure}
\label{app:dose}
A fraction of the target facts, drawn from outside the 300 evaluation facts, is written under the five request forms and the rest as statements, with statement-only earlier training; evaluation uses the 300 target facts written only as statements; at $100\%$ the token-matched stage writes 768 of the 1{,}000 target facts, including all 300 evaluation facts, in the five forms, so both columns remain measurements on unwritten forms. The declarative-paraphrase control writes the possessive and capital-city forms among its paraphrases; only the list and colon-ended declarative forms are scored. Exact generation, eight-seed means.

\begin{center}\small
\begin{tabular}{lcc}
\toprule
target facts in request forms & list & colon-ended declarative \\
\midrule
$0\%$ & 0.000 & 0.024 \\
$1\%$ & 0.122 & 0.508 \\
$5\%$ & 0.381 & 0.718 \\
$20\%$ & 0.744 & 0.880 \\
$100\%$ & 0.994 & 0.998 \\
$5\%$, declarative paraphrases instead & 0.003 & 0.062 \\
Pythia-1.4B, $20\%$ & 0.828 & -- \\
\bottomrule
\end{tabular}
\end{center}

\section{Replications}
\label{app:replications}

\paragraph{Independent fact worlds.} Two additional worlds were generated from independent generator seeds and run under the unchanged protocol. Earlier five-form training exceeds the statement-only condition on held-out list retrieval (pool top-1) by $+0.974$ and $+0.887$, $8/8$ seeds in each world; eight further seeds on the second world give $+0.972$, $8/8$.

\section{Generality}
\label{app:generality}
The separation, the dependence on the request's final words (keying), and the restore by rewrite or by the restoring vector, in every setting. Contrast columns: exact generation on the held-out list form after identical target-fact training, without and with earlier request-form training on unrelated facts. Keying: the statement-only model on \texttt{- \{Rel\} of \{e\} is} versus \texttt{The \{rel\} of \{e\}:}. Restore: the statement-only model's held-out list form, rewritten to end in ``is'' / with the model's own restoring vector (its mean declarative minus mean list context state) added at the answer position; the direction alone, used in Table~\ref{tab:reusable-direction}, gives $0.921$ in Pythia-410M. Eight seeds unless marked; Llama-3.2-1B four seeds. The five-relation rows are one training run on 1{,}000 facts across five relations and three entity types; declarative training-form accuracy is $0.97$--$1.00$ in every row. LoRA: rank 64, target stage only.

\begin{center}\small
\setlength{\tabcolsep}{4pt}
\begin{tabular}{lcc}
\toprule
& \multicolumn{2}{c}{list form} \\
\cmidrule(lr){2-3}
setting & statement-only & five-form \\
\midrule
Pythia-410M, capitals            & 0.000 & 0.977 \\
Pythia-1.4B                      & 0.040 & 0.997 \\
Qwen2.5-1.5B Base                & 0.603 & 0.997 \\
Llama-3.2-1B Base                & 0.048 & 0.997 \\
Five relations: capital          & 0.002 & 0.970 \\
\quad currency (country)         & 0.004 & 0.989 \\
\quad population (country)       & 0.038 & 0.991 \\
\quad founder (organization)     & 0.001 & 0.989 \\
\quad birth year (person)        & 0.344 & 0.982 \\
Birth years as target            & 0.150 & 0.952 \\
Shared-first-token capitals               & 0.001 & 0.982 \\
Targets written by LoRA          & 0.000 & 0.836 \\
\bottomrule
\end{tabular}
\end{center}

\begin{center}\small
\begin{tabular}{lcc}
\toprule
setting & keying (is / colon) & restore (base $\to$ rewrite / restoring vector) \\
\midrule
Pythia-410M, capitals & $0.948$ / $0.024$ & $0.000 \to 0.948$ / $0.945$ \\
Pythia-1.4B & $0.994$ / $0.266$ & $0.040 \to 0.994$ / $0.917$ \\
Qwen2.5-1.5B Base & $0.998$ / $0.482$ & $0.603 \to 0.998$ / $0.993$ \\
Llama-3.2-1B Base & $0.995$ / $0.125$ & $0.048 \to 0.995$ / $0.971$ \\
Birth years as target & $0.977$ / $0.221$ & $0.150 \to 0.977$ / $0.973$ \\
Shared-first-token capitals & $0.941$ / $0.025$ & $0.001 \to 0.941$ / $0.934$ \\
Targets written by LoRA & $0.708$ / $0.010$ & $0.000 \to 0.708$ / $0.730$ \\
\bottomrule
\end{tabular}
\end{center}

The five-relation run on Pythia-1.4B (eight seeds, list form, exact generation; every relation separates in $8/8$ seeds with matched declarative accuracy):

\begin{center}\small
\begin{tabular}{lcc}
\toprule
relation & statement-only & five-form \\
\midrule
capital & 0.119 & 0.998 \\
currency & 0.256 & 0.999 \\
population & 0.248 & 0.991 \\
founder & 0.089 & 0.996 \\
birth year & 0.685 & 0.996 \\
\bottomrule
\end{tabular}
\end{center}

The five relations learned in one run on Pythia-410M, keying (is / colon) and restore by rewrite:

\begin{center}\small
\begin{tabular}{lcc}
\toprule
relation & keying (is / colon) & restore by rewrite \\
\midrule
capital & $0.814$ / $0.200$ & $0.002 \to 0.814$ \\
currency & $0.730$ / $0.388$ & $0.004 \to 0.730$ \\
population & $0.960$ / $0.276$ & $0.038 \to 0.960$ \\
founder & $0.862$ / $0.299$ & $0.001 \to 0.862$ \\
birth year & $0.987$ / $0.522$ & $0.344 \to 0.987$ \\
\bottomrule
\end{tabular}
\end{center}

The birth-year statement-only list value ($0.344$) is higher than for the other relations: year answers come more readily after a colon; the five-form value is unaffected.

\section{Suffix Bank}
\label{app:suffix}
Sixteen request forms fixed before scoring, one to four per level of overlap with the training line \texttt{The capital of \{c\} is} (overlap = number of the training line's final tokens that recur immediately before the answer). Exact generation, eight-seed means, statement-only model.

\begin{center}\scriptsize\setlength{\tabcolsep}{4pt}
\begin{tabular}{cp{7.6cm}c}
\toprule
overlap & request form & generation \\
\midrule
0 & \texttt{- Capital of \{c\}:} & 0.000 \\
1 & \texttt{\{c\}'s capital is} & 0.039 \\
1 & \texttt{\{c\}'s capital city is} & 0.002 \\
1 & \texttt{For \{c\}, the capital is} & 0.022 \\
2 & \texttt{The capital city for \{c\} is} & 0.546 \\
2 & \texttt{The capital in \{c\} is} & 0.891 \\
2 & \texttt{The seat of government for \{c\} is} & 0.164 \\
2 & \texttt{The chief city in \{c\} is} & 0.085 \\
3 & \texttt{The capital city of \{c\} is} & 0.885 \\
3 & \texttt{- Capital of \{c\} is} & 0.948 \\
3 & \texttt{The seat of government of \{c\} is} & 0.359 \\
3 & \texttt{The main city of \{c\} is} & 0.187 \\
4 & \texttt{I know that the capital of \{c\} is} & 0.987 \\
4 & \texttt{The city that is the capital of \{c\} is} & 0.895 \\
4 & \texttt{Indeed, the capital of \{c\} is} & 0.995 \\
4 & \texttt{So the capital of \{c\} is} & 0.992 \\
\bottomrule
\end{tabular}
\end{center} 
\section{Patching Details}
\label{app:patching}

Target: statement-only model after target-fact training, list form, facts correct under the declarative form by pool top-1. Sites: last country token and answer position. Layers 0--6. Layer $\ell$ here is the residual stream entering block $\ell$; in the direction, shift, and linear-map analyses it is the output of block $\ell$. Sources: the same model's declarative answer-position state for the same fact; a norm-matched random vector; the same model's declarative state for a different fact; the cross-relation and entity-free sources are the same model's answer-position states under \texttt{The currency of \{c\} is}, \texttt{The population of \{c\} is}, and \texttt{The weather today is} (one state, broadcast to all items). The patched state is held at the prompt's final position on every decoding step; ranking is read from the first forward pass and generation from the held run.

\paragraph{Pythia-410M, full table.} Exact generation by layer (0--6) and source, eight-seed means; unpatched $0.000$; rewrite $0.948$. At layer 0 every non-random source is the final token and equals the rewrite; the layer-0 cells and the rewrite are scored on all 300 facts. Patching at the country token leaves generation at $0.000$ at every layer.

\begin{center}\small
\begin{tabular}{lccccccc}
\toprule
source & 0 & 1 & 2 & 3 & 4 & 5 & 6 \\
\midrule
same fact & 0.948 & 0.950 & 0.963 & 0.962 & 0.963 & 0.947 & 0.779 \\
different fact & 0.948 & 0.952 & 0.958 & 0.959 & 0.951 & 0.931 & 0.701 \\
currency & 0.948 & 0.949 & 0.961 & 0.960 & 0.946 & 0.934 & 0.757 \\
population & 0.948 & 0.951 & 0.959 & 0.953 & 0.928 & 0.920 & 0.735 \\
no entity & 0.948 & 0.927 & 0.895 & 0.834 & 0.734 & 0.543 & 0.018 \\
random & 0.000 & 0.000 & 0.000 & 0.000 & 0.000 & 0.000 & 0.000 \\
\bottomrule
\end{tabular}
\end{center}

\paragraph{Pythia-1.4B.} Same protocol on the Pythia-1.4B statement-only finals (eight seeds; on the facts correct under the declarative prompt, unpatched generation $0.040$; rewrite $0.995$). Exact generation by layer and source.

\begin{center}\small
\begin{tabular}{lccccccc}
\toprule
source & 0 & 1 & 2 & 3 & 4 & 5 & 6 \\
\midrule
same fact & 0.995 & 0.996 & 0.987 & 0.977 & 0.962 & 0.890 & 0.872 \\
different fact & 0.995 & 0.995 & 0.985 & 0.974 & 0.965 & 0.895 & 0.874 \\
currency & 0.995 & 0.995 & 0.988 & 0.983 & 0.971 & 0.926 & 0.928 \\
population & 0.995 & 0.995 & 0.988 & 0.982 & 0.972 & 0.923 & 0.908 \\
no entity & 0.995 & 0.995 & 0.948 & 0.916 & 0.723 & 0.448 & 0.140 \\
random & 0.000 & 0.000 & 0.000 & 0.000 & 0.000 & 0.000 & 0.000 \\
\bottomrule
\end{tabular}
\end{center}

\paragraph{Qwen2.5-1.5B.} Same protocol on the Qwen2.5-1.5B statement-only finals (eight seeds; unpatched generation $0.603$; rewrite $0.998$). Every ``is''-final source restores the list form at every layer 0--6 (same fact $0.998$--$0.995$, different fact $0.998$--$0.994$, currency $0.998$--$0.994$, population $0.998$--$0.994$, entity-free $0.998$--$0.989$); the norm-matched random state gives $0.000$ at every layer. The reverse intervention reduces declarative generation from $0.999$ to $0.550$; a norm-matched random removal leaves it at $0.999$.

\paragraph{Shared-first-token world.} Same protocol, on all 300 facts and with the same-fact, different-fact, and random sources, on the statement-only finals of the shared-first-token world (eight seeds; unpatched generation $0.001$; rewrite $0.941$). Exact generation of both capital tokens by layer and source.

\begin{center}\small
\begin{tabular}{lccccccc}
\toprule
source & 0 & 1 & 2 & 3 & 4 & 5 & 6 \\
\midrule
same fact & 0.941 & 0.941 & 0.943 & 0.945 & 0.932 & 0.910 & 0.724 \\
different fact & 0.941 & 0.942 & 0.938 & 0.936 & 0.922 & 0.892 & 0.633 \\
random & 0.000 & 0.000 & 0.000 & 0.000 & 0.000 & 0.000 & 0.000 \\
\bottomrule
\end{tabular}
\end{center}

\section{Context-State Intervention During Learning}
\label{app:aug}
From the statement-only stage-1 checkpoint, the byte-identical target stream is written for 1{,}200 updates while $d_f = \bar h^{f}_4 - \bar h^{\text{decl}}_4$ (the mean layer-4 context state of request form $f$ minus the mean declarative one, on the earlier-training facts; $f$ drawn per line from list, key--value, question, or none, with a seeded RNG) is added at the answer position of every target line at layer 4, moving the state toward the request form. For birth years, $d_f$ is computed on people 600--999 of the birth-year world (target-stream facts outside the evaluation set), in the capital-world stage-one model. Random: norm-matched Gaussian directions with the same draws. Held-out WikiText loss after training (Pythia-410M): targeted shift $5.584$, random shift $5.595$. Exact generation on $\evA$, eight-seed means (birth years: four seeds); the main-text values are the colon-ended declarative form, whose direction was not used in the shift.

\begin{center}\small
\begin{tabular}{llcccc}
\toprule
model & facts & form & no intervention & random shift & targeted shift \\
\midrule
Pythia-410M & capitals & declarative (training) & 0.995 & 0.997 & 0.995 \\
 & & colon-ended declarative & 0.024 & 0.100 & 0.943 \\
\addlinespace
Pythia-1.4B & capitals & declarative (training) & 0.995 & 0.997 & 0.997 \\
 & & colon-ended declarative & 0.266 & 0.332 & 0.904 \\
\addlinespace
Qwen2.5-1.5B & capitals & declarative (training) & 0.999 & 0.999 & 0.998 \\
 & & colon-ended declarative & 0.482 & 0.722 & 0.993 \\
\addlinespace
Pythia-410M & birth years & declarative (training) & 0.997 & 0.992 & 0.993 \\
 & & colon-ended declarative & 0.246 & 0.445 & 0.983 \\
\bottomrule
\end{tabular}
\end{center}

\section{Context-Direction Vector Details}
\label{app:rv}
Grid: $\ell\in\{4,6,8,10,12\}$, $\beta\in\{0.5,1.0,1.5,2.0\}$. Selection per seed on $A[300{:}600]$ by exact generation with the vector added at the request's answer position on every decoding step; the selected cell is recorded before $\evA$ is scored; the selected layer is 4 in every seed. Qwen2.5-1.5B fixes the layer at 4 and selects only $\beta$; Pythia-410M and Qwen2.5-1.5B break ties on pool top-1. The Appendix~\ref{app:generality} restoring vectors select layer and $\beta$ on this grid for every model, keeping the first best cell. The shared-first-token world's grid is offset one layer earlier. Fixed-norm removal: $u$ is projected out of $r$ and the result rescaled to $\|r\|$. The reverse intervention replaces the declarative answer-position state with the same fact's list-prompt state. Linear map: $M h^{f}_4(c)\approx h^{\text{decl}}_4(c)$ an affine map fit by ridge regression on the earlier-training facts $B[0{:}600]$ over the seven non-declarative evaluation forms, $\lambda$ chosen on $B[600{:}1000]$, applied at the answer position and held on every decoding step; on $\evA$: list $0.949$, declarative $0.995$; a rank-16 map gives the same values within $0.01$. Direction alone: $\|r\|\,\hat u$ added at the same site, $\beta$ chosen on $A[300{:}600]$ from $\{0.25,0.5,1,1.5,2,3\}$ ($\beta=1$ in every seed). Matched-cosine removals: a direction $v$ with $\cos(v,r)=\cos(u,r)$ is subtracted from $r$ and the result rescaled to $\|r\|$; the mirror of $u$ about $r$ has $\cos(v,u)=0.603$ and leaves $0.388$; ten random $v$ at that cosine have mean $\cos(v,u)=0.801$ and leave $0.151$; $u$ itself leaves $0.010$. In Qwen2.5-1.5B the random removal subtracts a norm-matched random direction and rescales to $\|r\|$. The punctuation direction $u_\ell$ used for the projection test is the difference of answer-position states at layer $\ell$ between \texttt{The value of \{c\} is} and \texttt{The value of \{c\}:}, averaged over the 300 evaluation country names; built instead from the birth-year world's people, the same direction has cosine $1.000$ with it; cosines with that layer's own list vector are $0.895$, $0.840$, $0.833$ at layers 4, 6, 8.

\section{Statistical Protocol}
\label{app:stats}

The training seed is the inferential unit. Core comparisons use eight paired seeds; regenerated worlds are analysed as separate eight-seed comparisons. Exact paired sign-flip tests enumerate all $2^8$ sign assignments; with eight same-signed differences the minimum two-sided $p$ is $0.0078125$; the $8/8$ counts reported in Appendices~\ref{app:replications} and~\ref{app:generality} are these tests. Confidence intervals use paired seed differences and the $t$ distribution with seven degrees of freedom, untruncated. Interventions applied to other request forms or during training carry a declarative-accuracy check: one that changes declarative accuracy by more than $0.05$ in any seed is reported as uninterpretable, never as a null; no such intervention exceeded it.

\end{document}